\documentclass{ecai} 

\usepackage{latexsym}
\usepackage{amssymb}
\usepackage{amsmath}
\usepackage{amsthm}
\usepackage{booktabs}
\usepackage{enumitem}
\usepackage{graphicx}
\usepackage{color}
\usepackage{algpseudocode}
\usepackage{multirow}
\usepackage{booktabs}
\usepackage[ruled,linesnumbered]{algorithm2e}

\newcommand{\BibTeX}{B\kern-.05em{\sc i\kern-.025em b}\kern-.08em\TeX}

\begin{document}


\begin{frontmatter}


\paperid{2842} 



\title{Enhancing Data Quality through Self-learning on Imbalanced Financial Risk Data}


\author[A]{\fnms{Xu}~\snm{Sun}} 
\author[A]{\fnms{Zixuan}~\snm{Qin}}
\author[A]{\fnms{Shun}~\snm{Zhang}}
\author[A]{\fnms{Yuexian}~\snm{Wang}}
\author[A]{\fnms{Li}~\snm{Huang}\thanks{Corresponding Author. Email: lihuang@swufe.edu.cn}}



\address[A]{School of Computing and Artificial Intelligence, Southwestern University of Finance and Economics}


\begin{abstract}
In the financial risk domain, particularly in credit default prediction and fraud detection, accurate identification of high-risk class instances is paramount, as their occurrence can have significant economic implications. Although machine learning models have gained widespread adoption for risk prediction, their performance is often hindered by the scarcity and diversity of high-quality data. This limitation stems from factors in datasets such as small risk sample sizes, high labeling costs, and severe class imbalance, which impede the models' ability to learn effectively and accurately forecast critical events. This study investigates data pre-processing techniques to enhance existing financial risk datasets by introducing TriEnhance, a straightforward technique that entails: (1) generating synthetic samples specifically tailored to the minority class, (2) filtering using binary feedback to refine samples, and (3) self-learning with pseudo-labels. Our experiments across six benchmark datasets reveal the efficacy of TriEnhance, with a notable focus on improving minority class calibration, a key factor for developing more robust financial risk prediction systems.
\end{abstract}

\end{frontmatter}


\section{Introduction}
\label{sec:label}
\begin{figure}[t]
    \centering
    \includegraphics[width=0.9\linewidth]{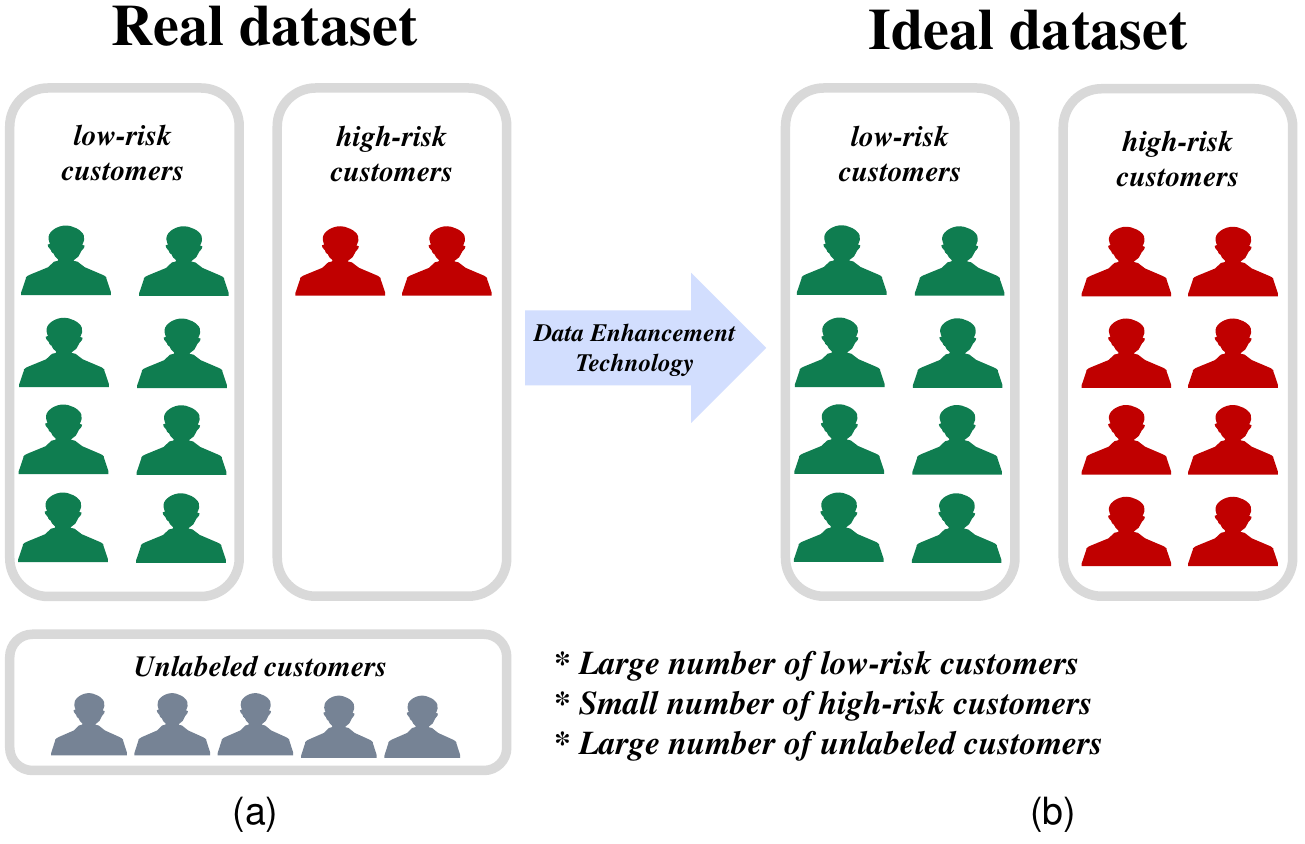}
    \caption{Ideal dataset vs Real dataset. (a) and (b) respectively represent the datasets in the real-world environment and the ideal environment. The data environment in reality faces challenges such as insufficient sample size, class imbalance, and a large amount of unlabeled data that has not been effectively utilized, while ideal data environment is characterized by sufficient sample size and balanced classes.}
    \label{fig:Ideal dataset vs Real dataset}
\end{figure}

A crucial aspect of managing individual customers in current financial institutions revolves around customer financial risk management, which primarily encompasses two categories of risk events: credit default and credit fraud. Credit default occurs when clients cannot repay loans, leading to financial institutions incurring losses on unpaid loan principal and interest~\citep{Nigmonov2021COVID19PR}. On the other hand, credit fraud involves unauthorized transactions such as credit card fraud, identity theft, and account takeovers. These not only lead to direct economic losses for both clients and financial institutions but also significantly impact client loyalty towards the institutions and can generate negative publicity, causing multiple losses. Therefore, it is essential to develop effective customer financial risk management models to mitigate the impact of such risk events on financial institutions. However, whether relying on statistical models or modern deep learning models, there is a significant reliance on high-quality data, while they faced three main challenges: insufficient samples~\citep{Chawla2004EditorialSI}, class imbalance~\citep{Krawczyk2016LearningFI}, and high labeling costs~\citep{Settles2009ActiveLL}, as shown in Figure~\ref{fig:Ideal dataset vs Real dataset}. In particular, the issue of class imbalance is critical, as the losses from a single negative case often outweigh the profits from several positive cases~\citep{Benbouzid2022BankCR,BORIO2014182,Nigmonov2021COVID19PR}. Consequently, addressing class imbalance is a fundamental concern in customer financial risk management.


Among these, altering the data distribution stands as the most versatile approach, encompassing techniques such as resampling, Generative Adversarial Networks (GANs), data filtering, and pseudo-labeling. Resampling techniques, further divided into oversampling and undersampling, aim to balance the dataset. Oversampling methods like Synthetic Minority Over-sampling Technique (SMOTE)~\citep{Chawla2002SMOTESM,Fernandez2018acf} generate synthetic instances to enhance model sensitivity towards minority classes. Undersampling techniques, exemplified by EasyEnsemble \citep{2009Exploratory}, reduce majority class samples to achieve balance. GANs, like those in ~\citep{Douzas2018EffectiveDG,Goodfellow2014GANs,mirza2014conditional,Xu2019ModelingTD}, generate high-quality synthetic data to improve minority class predictions by mimicking real data distributions. Data filtering, as discussed in ~\citep{Sez2016EvaluatingTC,Smith2011ImprovingCA,Wilson2000ReductionTF}, enhances dataset quality by removing noisy or less informative samples. Pseudo-labeling, a semi-supervised learning technique~\citep{Lee2013PseudoLabelT}, addresses label scarcity and labeling costs by generating and utilizing pseudo-labels for unlabeled samples, thereby expanding the dataset and improving model generalization.

While these techniques can enhance minority class recognition, they are not without limitations. Resampling techniques, such as those discussed in~\citep{Krawczyk2016LearningFI}, can result in information loss or overfitting. GAN-based approaches, as mentioned in~\citep{Arjovsky2017TowardsPM}, may lead to overfitting on synthetic samples, potentially overlooking the complexity of real data. Pseudo-labeling, as described in~\citep{Lee2013PseudoLabelT,Arazo2019PseudoLabelingAC}, can expand the training set, but poor pseudo-label quality or inappropriate generation strategies can introduce noise, thereby reducing model accuracy. Furthermore, these methods are often employed in isolation, without synergistic integration, which hampers their effectiveness in practical scenarios.

This paper focuses on the issue of class imbalance in financial risk data and proposes a method to enhance the quality of imbalanced datasets through self-learning, referred to as {\bf TriEnhance}. It effectively transforms the original dataset into a higher-quality dataset by integrating data synthesis, data filtering, and pseudo-label-based self-learning strategies. The method mainly includes the following contributions: 


\begin{itemize}
    \item We introduce the TriEnhance data enhancement method specifically designed for the financial risk domain. This method not only improves data quality and expands the dataset through data synthesis, data filtering, and pseudo-label self-learning strategies but also emphasizes generating high-quality data that reflect complex real-world financial behavior patterns, thereby significantly enhancing the model's predictive ability for key minority class events in financial risk.
    
    \item We introduce a novel data-processing algorithm, KFULF, which implements a K-fold pseudo-label strategy based on unknown label filtering to tap into the potential value of unlabeled data. By introducing "pseudo-labels" to mark samples where the model is uncertain, we effectively filter out samples with low model confidence while retaining high-confidence samples and their pseudo-labels, thus optimizing the structure of the training set.

    \item Through extensive experiments, we validate the effectiveness of our method in enhancing multiple financial risk datasets, thereby improving the performance of existing models. These experimental results confirm that TriEnhance is not only effective but also widely applicable.
\end{itemize}


\section{Related work}
\label{sec:Related work}
\begin{itemize}

    \item \textbf{Resampling Techniques:}
    Resampling techniques are fundamental in addressing class imbalances, particularly in financial risk management. The Synthetic Minority Over-sampling Technique (SMOTE) introduced by Chawla \emph{et al.} is among the most influential methods\citep{Chawla2002SMOTESM}, generating synthetic examples by interpolating between existing minority class instances. This technique has spurred numerous variants, each aiming to refine the approach to better capture the underlying distribution of minority class data. Another method, Adaptive Synthetic Sampling (ADASYN)~\citep{He2008ADASYNAS}, adjusts the synthetic sample generation by using a density distribution, as detailed by He \emph{et al.}, to provide better adaptive behavior in imbalanced learning scenarios. Additionally, undersampling techniques like EasyEnsemble\citep{2009Exploratory}, also developed by He \emph{et al.}, selectively reduce the number of majority class samples, thereby providing a balanced dataset that enhances the model’s sensitivity to the minority class.

    \item \textbf{Generative Adversarial Networks:}
    The application of GANs for synthetic data generation in imbalanced datasets has been explored extensively, with innovations aiming to generate realistic samples that can help models learn more effective decision boundaries. Goodfellow \emph{et al.} introduced GANs\citep{Goodfellow2014GANs}, which have been adapted for various tasks including financial fraud detection. Conditional GANs (CGANs)\citep{mirza2014conditional}, introduced by Mirza \emph{et al.}, further allow for the generation of samples conditioned on class labels, enabling more targeted synthesis of minority class examples. Following these, the introduction of Wasserstein GANs (WGANs) by Arjovsky \emph{et al.} marked a significant improvement in the stability of GAN training\citep{arjovsky2017wasserstein}, offering better convergence and handling of diverse data distributions, a key advantage in financial applications where model stability is crucial. Another significant advancement in this area is the Conditional Tabular GAN (CTGAN) introduced by Xu \emph{et al.}\citep{Xu2019ModelingTD}, designed specifically to handle the unique challenges of tabular data with mixed data types and skewed distributions. CTGAN enhances the diversity and representativeness of training datasets, particularly useful in financial risk scenarios where data often have complex.

    \item \textbf{Data Filtering Techniques:}
    Data filtering techniques aim to improve the quality of training datasets by selectively removing noisy or irrelevant samples that may degrade learning performance. Wilson's Editing \citep{Wilson1972AsymptoticPO}, one of the pioneering techniques in this area, employs nearest neighbor rules to detect and eliminate mislabeled or noisy data, thereby enhancing the overall quality of the training set for better classifier performance. This method is particularly effective in contexts where the data may contain human errors or artifacts in the data collection process.In addition to traditional approaches, recent developments have focused on more sophisticated integrations of machine learning algorithms with data filtering techniques. These modern approaches aim to adaptively identify and remove outliers and noise, making them highly effective even in dynamic and complex data environments. For instance, Japkowicz et al. \citep{Japkowicz1995AND} explored an iterative filtering process that significantly enhances the classifier's performance on noisy datasets by continuously refining the training set through repeated cycles of training and filtering. This method has shown promise in applications ranging from fraud detection to automated quality control systems, where maintaining data integrity is crucial.

    \item \textbf{Pseudo-labeling Techniques:}
    Pseudo-labeling\citep{Lee2013PseudoLabelT}, a semi-supervised learning approach, utilizes unlabeled data by assigning temporary labels based on the predictions of a trained model. This technique, first discussed by Lee in the context of deep neural networks, has been extensively applied to enhance model training where labeled data are scarce. The approach involves using the confident predictions of a model to generate labels for unlabeled data, which are then used to retrain the model, progressively improving its accuracy on both labeled and unlabeled datasets.
    
\end{itemize}


\section{Methodologies}
\label{sec:method}
\begin{figure*}[h]
    \centering
    \includegraphics[width=\textwidth]{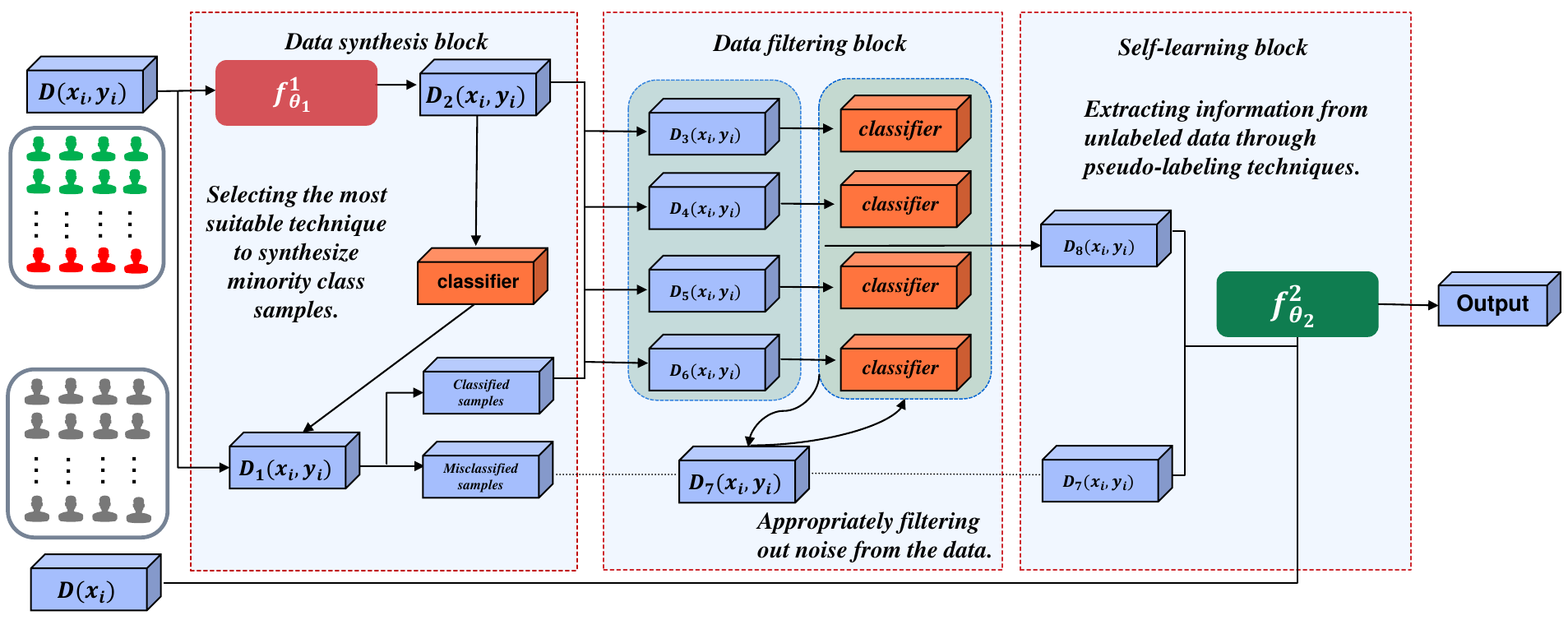}
    \caption{Overview of the TriEnhance Architecture.}
    \label{fig:Architecture diagram}
    \vspace{-0.4cm}
\end{figure*}

To address the challenges of accurately predicting minority class samples in imbalanced data settings, we introduce the TriEnhance, a data enhancement framework, as shown in Figure~\ref{fig:Architecture diagram}. This comprehensive strategy enhances the model's ability to detect minority class occurrences by incorporating data synthesis, dynamic data filtering, and innovative self-learning techniques. Initially, the data synthesis block utilizes an adaptive algorithm to assess techniques like SMOTE and CTGAN based on the F1 score, facilitating automatic selection to enhance the representation and diversity of minority categories. Subsequently, the data filtering block dynamically adjusts the difficulty threshold based on supervisory signals from the data synthesis block, effectively eliminating noise and challenging samples. Lastly, the self-learning block leverages unlabeled data through the {\bf K}-{\bf F}old {\bf U}nknown-{\bf l}abel {\bf F}iltering algorithm, KFULF, a novel approach for filtering unknown labels using a K-fold pseudo-label method and a pseudo-label strategy guided by sample confidence ranking.


\subsection{Data Synthesis Block}
\begin{figure}
    \centering
    \includegraphics[width=0.9\linewidth]{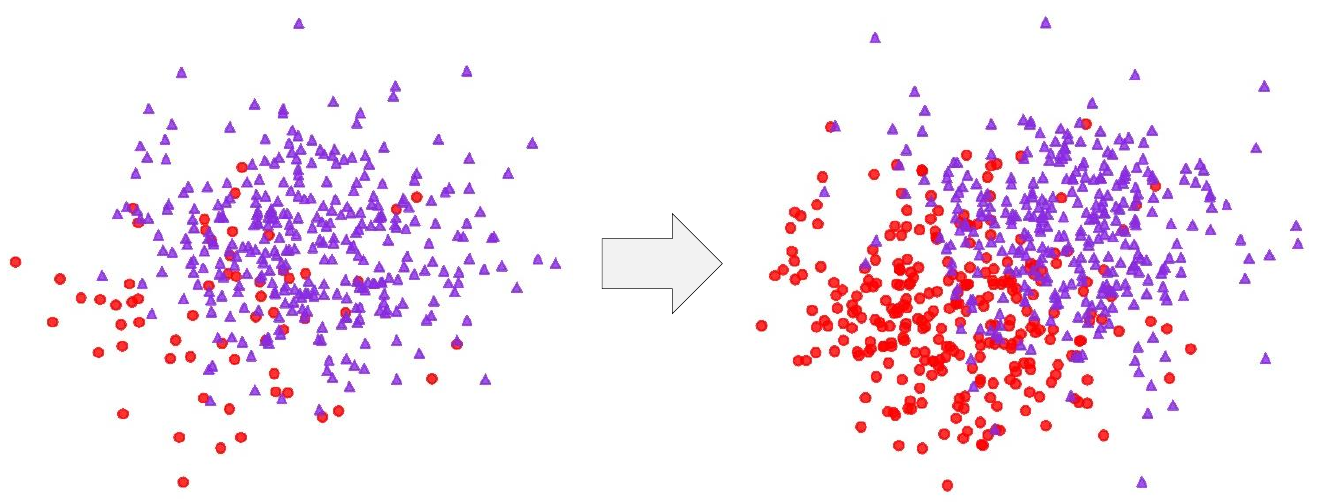}
    \caption{Comparison of distribution before and after data synthesis.}
    \label{fig:Comparison of distribution before and after data synthesis}
    \vspace{-0.6cm}
\end{figure}

The data synthesis block aims to enhance the model's ability to recognize minority classes in imbalanced datasets by increasing the number of minority class samples. Specifically, this module employs an adaptive algorithm based on the F1 score to evaluate various data synthesis techniques, including SMOTE and CTGAN. This enables the algorithm to automatically select the best-performing technique, ensuring that TriEnhance can choose the optimal synthesis strategy for various financial risk datasets. This method improves the representation of minority classes, ultimately enhancing the model's prediction accuracy for these classes. The distribution of the original data after processing through the Data Synthesis Block is shown in Figure~\ref{fig:Comparison of distribution before and after data synthesis}.

\subsubsection{Data Synthesis}
We consider a binary classification problem with the output variable $Y \in \{0, 1\}$, where $Y=1$ represents the minority class and $Y=0$ represents the majority class. The prior probabilities of the classes are denoted as $p(y=1) = p_1$ and $p(y=0) = p_0$ with $p_1 \ll p_0$, indicating a significant class imbalance. The classifier $f(x)$ predicts the class of an input $x$. We define the classifier's decision function, $f$, as follows:
\begin{equation}
    f(x) = \begin{cases} 
    1 & \text{if } p(y=1 \mid x) \geq t, \\
    0 & \text{otherwise},
    \end{cases}
\end{equation}
where $t$ is a decision threshold, typically set to $0.5$. The expected risk associated with the classifier, denoted $R(f)$, under a 0-1 loss function, which can be calculated by:
\begin{equation}
    R(f) = E[\mathcal{L}(f(x), y)].
\end{equation}
Here, the loss function $\mathcal{L}(f(x), y)$ employs the Kronecker delta function $\delta$, defined as:
\begin{equation}
    \mathcal{L}(f(x), y) = 1 - \delta(f(x), y),
\end{equation}
where $\delta(f(x), y) = 1$ if $f(x) = y$ and $0$ otherwise. This function measures the discrepancy between predictions and actual labels.

Expanding the expected risk, we express it as:
\begin{equation}
    R(f) = p_1 \cdot \Pr \big(f(x) = 0 \mid y=1 \big) + p_0 \cdot \Pr \big(f(x) = 1 \mid y=0 \big),
\end{equation}
which considers the weighted sum of misclassification probabilities, it is essential to account for both false negatives and false positives. By introducing supplementary samples from the minority class, we are able to modify the prior probabilities. Specifically, the adjusted probability for the minority class, denoted as $p_1'$, increases from $p_1$ to $p_1' > p_1$, while the adjusted probability for the majority class, denoted as $p_0'$, decreases from $p_0$ to $p_0' < p_0$. Subsequently, this adjustment results in a change in the anticipated risk to:
\begin{equation}
    R'(f) = p_1' \cdot \Pr(f(x) = 0 \mid y=1) + p_0' \cdot \Pr(f(x) = 1 \mid y=0).
\end{equation}
This adjustment amplifies the penalty for misclassifying the minority class, leading to a higher cost for false negatives, while diminishing the penalty for misclassifying the majority class, thereby reducing the impact of false positives. As a result, the classifier is encouraged to enhance its sensitivity towards the minority class, thereby improving its recall.

\subsubsection{Meta-Synthetic-Data Learning}
Meta-synthetic-data learning initially partitions the dataset $D$ into a sub-training set and validation set, i.e., $D = {D^{tr} \cup D^{val}}$. For each predefined data synthesis technique, we utilize $D^{tr}$ to generate synthetic data $D^{syn}$, which is then merged with the sub-training set data to form an augmented dataset $D_{\text{aug}}$. Subsequently, we train a model $ M $ on this augmented dataset and evaluate it on the validation set $D_{\text{val}}$ to obtain the F1 scores for each technique. After systematically evaluating all candidate techniques, we select the technique with the highest F1 score. Subsequently, we integrate the samples correctly classified by model $ M $ on the validation set $ D_{\text{val}} $ into the augmented dataset $ D_{\text{aug}} $ generated by this technique, updating $ D_{\text{aug}} $. Finally, we return $ D_{\text{aug}} $ along with the misclassified samples $ D_{\text{mis}} $  from the validation set. $ D_{\text{mis}} $ will be utilized for further data enhancement in the data filtering block of the TriEnhance method. Pseudocode can be found in Algorithm~\ref{ag-meta}.
The algorithm's core is the dynamic selection of the optimal data synthesis technique, like SMOTE or CTGAN. This performance-based approach not only leverages existing technologies but also adapts to future advancements, allowing TriEnhance to adjust to real-world data variability.

\begin{algorithm}
\caption{Meta-Synthetic-Data Learning}
\label{ag-meta}
\KwIn{Dataset $D$, List of Data Synthesis Techniques $T$}
\KwOut{$D_{\text{aug}}$, $D_{\text{mis}}$}

$D^{tr}$, $D^{val}$ ← Partition($D$). \\

$F1_{\text{sc}}$ ← [ ] 

\For{$t$ in $T$} {
    $\text{Generate minority class data.}$ \\
    $D^{syn}$ ← $t$.Generate($D^{tr}$, minority=True). \\ 
    $D_{\text{aug}}$ ← $D^{tr} \cup D^{syn}$. \\
    $M$ ← Train($D_{\text{aug}}$). \\
    $F1_{t}$ ← Evaluate($M$, $D^{val}$). \\
    $F1_{\text{sc}}$.append($F1_{t}$)  
}

$\text{best\_tech}$ ← ArgMax($F1_{\text{sc}}$).  \\ 
$\text{Generate minority class data.}$ \\
$D^{syn}$ ← $\text{best\_tech}$.Generate($D^{tr}$, minority=True). \\ 
$D_{\text{aug}}$ ← $D^{tr} \cup D^{syn}$.  \\
$M$ ← Train($D_{\text{aug}}$).  \\

$D_{\text{aug}}$ ← $D_{\text{aug}} \cup$ Correct($D^{val}$). \\
$D_{\text{mis}}$ ← Misclassified($D^{val}$). \\

Return $D_{\text{aug}}$, $D_{\text{mis}}$. 
\end{algorithm}
\vspace{-0.6cm}

\subsection{Data filtering Block}
\begin{figure}[h]
    \centering
    \includegraphics[width=0.9\linewidth]{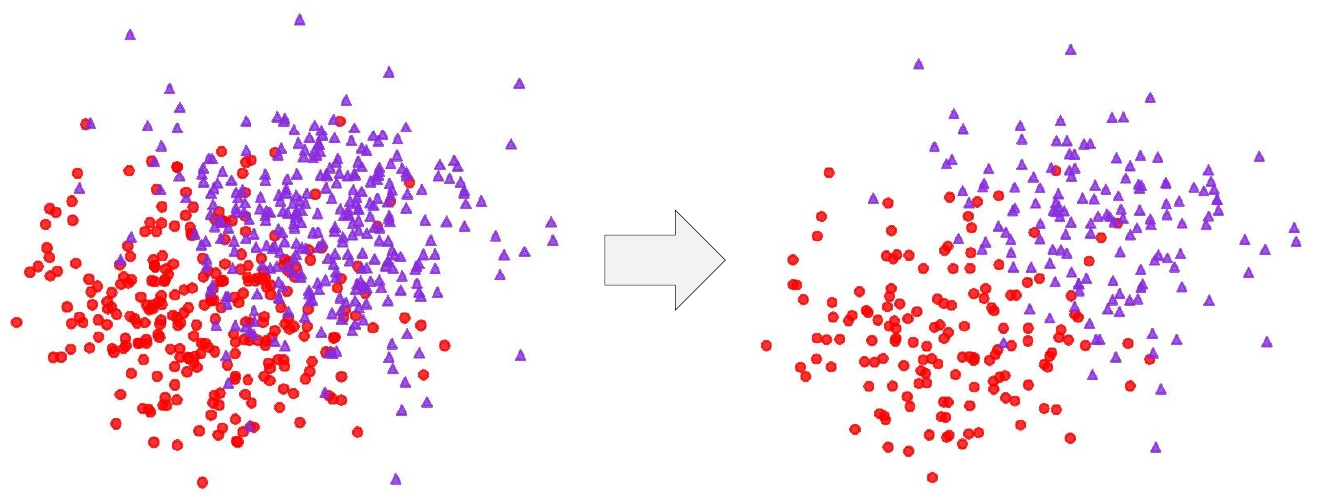}
    \caption{Comparison of distribution before and after data filtering.}
    \label{fig:Comparison of distribution before and after data filtering}
    \vspace{-0.1cm}
\end{figure}

Techniques like SMOTE and CTGAN can augment samples from minority classes to mitigate class imbalance. However, these methods may introduce noise and outliers that are challenging to classify, then disrupt decision boundaries and diminish model learning efficiency and predictive accuracy. To tackle this challenge, we introduce a data filtering method that dynamically adjusts difficulty thresholds to filter out noisy and hard-to-classify samples, associated with a sample retention strategy based on the original data's class proportions to ensure model robustness. The distribution of data after passing through the Data filtering Block is shown in Figure~\ref{fig:Comparison of distribution before and after data filtering}.


Define the model's prediction function as $\hat{f}$ and the prediction error of the model's predictions as:
\begin{equation}
     \text{Err}(x) = E[(Y - \hat{f}(x))^2],
\end{equation}
which can be further decomposed into:
\begin{equation}
    \text{Err}(x) = (\text{Bias}[\hat{f}(x)])^2 + \text{Var}[\hat{f}(x)] + \sigma^2.
\label{eq:Err_x}
\end{equation}
Herein:
\begin{itemize}
    \item $\text{Bias}[\hat{f}(x)]$: $E[\hat{f}(x)] - f(x)$, represents the difference between the expected prediction of the model and the true value.
    \item $\text{Var}[\hat{f}(x)]$: $E[(\hat{f}(x) - E[\hat{f}(x)])^2]$, describes the variability of model predictions for different training datasets.
    \item $\sigma^2$: determined by the noise inherent in the data, assumed to be constant.
\end{itemize}

Considering a dataset within noise, predictions for specific data points $x$ can exhibit high variance as the model attempts to fit these outliers. Let $p_{\text{noisy}}$ be the probability that a point is noisy, where these points should have a higher prediction variance than normal:
\begin{equation}
    \text{Var}_{\text{noisy}}[\hat{f}(x)] > \text{Var}_{\text{normal}}[\hat{f}(x)].
\end{equation}
Thus, filtering out these noise points can reduce the average variance of the model:
\begin{equation}
    \text{Var}_{\text{new}}[\hat{f}(x)] < \text{Var}_{\text{old}}[\hat{f}(x)].
\end{equation}
This occurs because the model no longer needs to accommodate data points that lead to high predictive uncertainty. Removing noise points might slightly increase the model's bias since the model becomes simplified and may not capture some complex patterns. Let $\Delta \text{Bias}$ be the increase in bias, which may not exceed the difference in bias before and after filtering:
\begin{equation}
    \Delta \text{Bias} \approx \text{Bias}_{\text{new}}[\hat{f}(x)] - \text{Bias}_{\text{old}}[\hat{f}(x)].
\end{equation}
Typically, $\Delta \text{Bias}$ is small because the increase in bias is usually much less significant than the reduction in variance.

Combining significant variance reduction and a slight increase in bias, the overall prediction error may decrease:
\begin{equation}
    \text{Err}_{\text{new}}(x) = (\text{Bias}_{\text{new}}[\hat{f}(x)])^2 + \text{Var}_{\text{new}}[\hat{f}(x)] + \sigma^2 < \text{Err}_{\text{old}}(x).
\end{equation}
This demonstrates that filtering out noise and hard-to-classify points, although slightly increasing bias, reduces the overall error due to a significant decrease in variance, thus improving the model's generalization ability on unseen data.

The data filtering block follows the data synthesis block. By analyzing the model $M$'s prediction probabilities on the augmented dataset $D_{\text{aug}}$, we calculate the difference between the highest probability $p_{\text{max}}$ and the second-highest probability $p_{\text{sec\_max}}$ for each sample, denoted as $ \Delta p = p_{\text{max}} - p_{\text{sec\_max}}$. Based on this difference relative to a preset difficulty threshold $t$, we determine whether to retain specific samples. Subsequently, we retrain the model $M$ on various $D_{\text{filtered}}$ datasets derived under different thresholds $t$ and calculate the F1 score on $D_{\text{mis}}$. We select the $D_{\text{filtered}}$ corresponding to the highest F1 score. This approach allows us to dynamically adjust the difficulty threshold $t$ to optimize filtering effectiveness, effectively eliminating noise and hard-to-classify samples.

To avoid information loss and overfitting from excessive filtering, we employ a sample retention strategy that is based on the class proportions of the original dataset. For each class, we determine the number of samples to retain as follows: \( n(c) = p(c) \times N_{\text{filtered\_out}} \), where \( N_{\text{filtered\_out}} \) represents the total count of filtered-out samples. The retained samples are then reintegrated into \( D_{\text{filtered}} \). This strategy helps maintain dataset diversity and class ratios.

\subsection{Self-learning block}



\begin{figure}[h]
    \centering
    \includegraphics[width=0.9\linewidth]{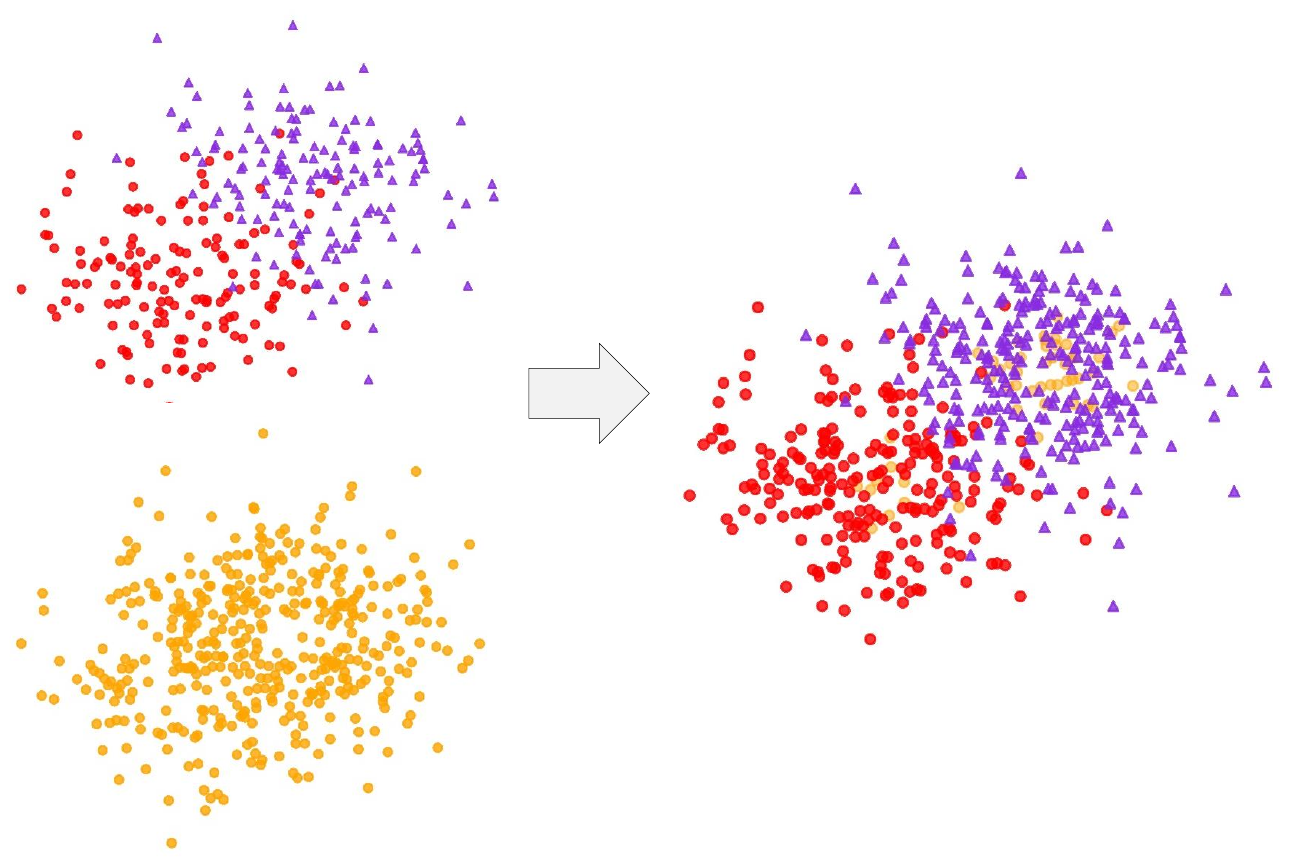}
    \caption{Comparison of distribution of data before and after self-learning.}
    \label{fig:Comparison of distribution of data before and after self-learning}
\end{figure}

Unlabeled data form the predominant portion of real-world datasets and possess significant value. However, data generation is an ongoing process, and the expense of manual data annotation is excessively high. Therefore, developing a self-learning module to investigate the importance of unlabeled data holds paramount importance. The distribution of the data after passing through the self-learning block is shown in Figure~\ref{fig:Comparison of distribution of data before and after self-learning}. 

Considering pseudo-labeled sample set $D_{\text{pseudo}}$, the re-construct training dataset enhanced as: $D_{\text{enhanced}} = D \cup D_{\text{pseudo}}$. Within $D_{\text{enhanced}}$, due to the inclusion of more high-confidence samples that equation closely with the true distribution, variance is reduced:
\begin{equation}
    \text{Var}_{\text{enhanced}}[\hat{f}(x)] < \text{Var}_{\text{original}}[\hat{f}(x)].
\end{equation}
This indicates that the model adapts more stably to the training data, reducing prediction fluctuations caused by randomness in the training data. The introduced data points are not only high in confidence but are also likely to be close to their true class labels, hence the impact on bias is minimal. While the model may become slightly more complex due to the addition of accurate information, bias may increase slightly:
\begin{equation}
    \Delta \text{Bias} \approx \text{Bias}_{\text{enhanced}}[\hat{f}(x)] - \text{Bias}_{\text{original}}[\hat{f}(x)]
\end{equation}
Typically, $\Delta \text{Bias}$ is small because the increase in bias is much less significant compared to the reduction in variance. Therefore, by adding high-confidence pseudo-labeled samples, the overall expected model error decreases:
    \begin{equation}
        \begin{aligned}
            \text{Err}_{\text{enhanced}}(x) & = \text{Bias}^2_{\text{enhanced}}[\hat{f}(x)] \\
                                           & + \text{Var}_{\text{enhanced}}[\hat{f}(x)] + \sigma^2 \\
                                           & < \text{Err}_{\text{original}}(x)
        \end{aligned}
    \end{equation}

Following this concept, we propose a task-related incremental enhancement self-learning block that continuously incorporates high-confidence pseudo-labeled samples into the training dataset, including the K-fold Unknown-label Filtering and the Pseudo-label Generation Strategy.

\noindent{\bf K-Fold Unknown-label Filtering~(KFULF).} To enhance the utilization of unlabeled data, KFULF employs a filtering mechanism that selects high-confidence pseudo-labeled samples. In each iteration, the unlabeled data is initially divided into $N$ subsets, with $N-1$ subsets being assigned artificial label~(e.g., ``$-1$") and the remaining subset designated as the test set. Subsequently, these artificially labeled datasets are combined to form the training data for retraining. Following this, the well-trained model predicts the unlabeled test set, identifying samples with explicit predictions as high-confidence samples. Upon completion of the K-Fold cycle, all high-confidence samples are incorporated into the training set.

\begin{algorithm}
\caption{K-Fold Unknown-label Filtering (KFULF)}
\label{alg:KFULF}
\KwIn{Training features $X_{tr}$, Training labels $y_{tr}$, Unlabeled data $X_{ul}$, Classifier $M$, Pseudo-label $-1$}
\KwOut{$X_{en}$, $y_{en}$}

$X^{ul}_k$ ← Split($X_{ul}$, $K$). \\

$D_{mk}$ ← [ ]  \\
$L$ ← [ ]  \\

\For{$k = 1$ \textbf{to} $K$} {
    $X^{tr}_u$ ← Concat($X_{ul}$, except=$k$). \\  
    $X^{te}_u$ ← $X^{ul}_k$. \\  
    
    $X^{tr}_c$ ← $X_{tr} \cup X^{tr}_u$. \\  
    $y^{tr}_c$ ← $y_{tr} \cup$ [-1] * $len(X^{tr}_u)$. \\  
    
    $M$ ← Train($X^{tr}_c$, $y^{tr}_c$). \\
    
    $y_{pred}$ ← $M$.predict($X^{te}_u$). \\  
    
    $X_{ft}$ ← Filter($X^{te}_u$, condition=$y_{pred} \neq -1$). \\  
    $y_{ft}$ ← Filter($y_{pred}$, condition=$y_{pred} \neq -1$). \\  
    
    $D_{mk}$ ← Append($D_{mk}$, $X_{ft}$). \\  
    $L$ ← Append($L$, $y_{ft}$). \\  
}

$X_{en}$ ← $X_{tr} \cup D_{mk}$. \\
$y_{en}$ ← $y_{tr} \cup L$. \\

\Return $X_{en}$, $y_{en}$. \\
\end{algorithm}

\noindent{\bf Delay-decision Strategy~(DDS).} Inspired by human decision-making, which often involves delaying decisions in times of uncertainty, we introduce a delay-decision strategy. This strategy iteratively assesses unlabeled samples and incorporates the top $30\%$ highest confidence samples as high-confidence pseudo-labeled samples into the training set, those remaining samples are waiting for the next time precision. After the iteration, if there is no further improvement in the model's performance or all unlabeled data has been utilized, the algorithm reaches its conclusion.

\begin{algorithm}
\caption{Delay-decision Strategy (DDS)}
\label{alg:DDS}
\KwIn{Training features $X_{tr}$, Training labels $y_{tr}$, Unlabeled data $X_{ul}$, Classifier $M$, Target percentage $tgt\_pct$}
\KwOut{$X_{tr\_fin}$, $y_{tr\_fin}$}

$impv $ ← $ True$,
$F1_{\text{base}}$ ← $ CalcF1(M, X_{tr}, y_{tr})$

$X_{en}$ ← [ ],
$y_{en}$ ← [ ] 

\While{$impv$ \textbf{and} $len(X_{ul}) > 0$} {
    $y\_proba $ ← $ M.\text{PredictProba}(X_{ul})$ \\
    $top\_idx $ ← $ \text{SelectTop}(\text{SortBy}(y\_proba), pct=tgt\_pct)$ \\
    $X_{sel} $ ← $ X_{ul}[top\_idx]$, 
    $y_{sel} $ ← $ M.\text{Predict}(X_{sel})$ \\
    
    $X_{tmp} $ ← $ \text{Concat}(X_{tr}, X_{en}, X_{sel})$ \\
    $y_{tmp} $ ← $ \text{Concat}(y_{tr}, y_{en}, y_{sel})$ \\

    $M $ ← $ \text{Retrain}(X_{tmp}, y_{tmp})$ \\
    $F1_{\text{new}}$ ← $ \text{CalcF1}(M, X_{tmp}, y_{tmp})$ 

    \If{$F1_{\text{new}}$ > $F1_{\text{base}}$} {
        $F1_{\text{base}}$ ← $ F1_{\text{new}}$,  
        $X_{en} $ ← $ \text{Concat}(X_{en}, X_{sel})$ \\
        $y_{en} $ ← $ \text{Concat}(y_{en}, y_{sel})$ \\
        $X_{ul} $ ← $ \text{Drop}(X_{ul}, top\_idx)$  \\
    } \Else {
        $impv $ ← $ False$  
    }
}

$X_{tr\_fin} $ ← $ \text{Concat}(X_{tr}, X_{en})$  
$y_{tr\_fin} $ ← $ \text{Concat}(y_{tr}, y_{en})$  

\Return $X_{tr\_fin}$, $y_{tr\_fin}$
\end{algorithm}


When TriEnhance working, the self-learning block will adaptively select the most suitable pseudo-labeling techniques from KFULF and DDS for the current dataset. It effectively selects valuable samples from unlabeled data to continuously enhance the quality of the training dataset.

\section{Experimental}
\label{sec:experiments}
\subsection{Datasets and preprocessing}


We utilized six financial risk datasets from Kaggle and the UCI Machine Learning Repository to validate the effectiveness of TriEnhance. The basic statistical information of the datasets is listed in Table \ref{table:dataset_summary_ir}, where IR represents the imbalance ratio of each dataset. To ensure the validation of TriEnhance's effectiveness under a uniform standard, we applied consistent preprocessing steps to all datasets: (1) removing columns with more than $50\%$ missing data and filling the remaining missing values with the mode of each respective column; (2) encoding non-numeric features with label encoding.

\begin{itemize}

    \item \textbf{Bank Loan Status Dataset \footnote{https://www.kaggle.com/datasets/zaurbegiev/my-dataset/data/} (BLSD)}: Sourced from Kaggle, it comprises 100,515 samples across 18 features, focusing on loan default prediction.

    \item \textbf{Taiwan Credit Dataset \footnote{https://archive.ics.uci.edu/ml/datasets/default+of+credit+card+clients/} (TCD)}: Sourced from the UCI Machine Learning Repository, this dataset includes 30,000 client records with 23 features, contributed by Chung Hua University and Tamkang University in 2016. 

    \item \textbf{Zhongyuan Credit Dataset \footnote{https://www.datafountain.cn/competitions/530/datasets} (ZCD)}: Provided by the 2021 CCF competition, this dataset contains 10,000 loan history records, including loan records, user information, occupation, age, and marital status.

    \item \textbf{Give Me Some Credit \footnote{https://www.kaggle.com/competitions/GiveMeSomeCredit/} (GMSC)}: Comprising 150,000 records with 10 features, this Kaggle dataset is aimed at forecasting potential default risk.
    
    \item \textbf{Credit Card Fraud Detection \footnote{https://www.kaggle.com/mlg-ulb/creditcardfraud/} (CCFD)}: Obtain from Kaggle, this dataset includes 284,807 transactions made by European cardholders in September 2013, with 492 being frauds. Due to confidentiality, features are PCA-transformed numerical variables, except for 'Time' and 'Amount'. 
    
    \item \textbf{Synthetic Financial Datasets For Fraud Detection \footnote{https://www.kaggle.com/datasets/ealaxi/paysim1/data/} (SFDFD)}: A synthetic dataset from Kaggle, generated by the PaySim simulator based on real transaction samples from mobile financial services in an African country. 

\end{itemize}

\begin{table}[h]
    \caption{Basic statistical information on financial risk datasets.}
    \centering
    \footnotesize
    \begin{tabular}{lcccccc} 
    \toprule
    Dataset & Samples & Features & Pos Ratio & Neg Ratio & IR \\
    \midrule
    BLSD & 100,515 & 18 & 0.2252 & 0.7748 & 1:3.44 \\
    TCD & 30,000 & 23 & 0.2212 & 0.7788 & 1:3.52 \\
    ZCD & 10,000 & 38 & 0.1683 & 0.8317 & 1:4.94 \\
    GMSC & 150,000 & 10 & 0.0668 & 0.9332 & 1:13.97 \\
    CCFD & 284,807 & 30 & 0.0017 & 0.9983 & 1:587.65 \\
    SFDFD & 6,362,620 & 10 & 0.0013 & 0.9987 & 1:768.23 \\
    \bottomrule
    \end{tabular}
    \label{table:dataset_summary_ir}
    \vspace{-0.2cm}
\end{table}

\subsection{Metrics}
To comprehensively assess the effectiveness of TriEnhance in enhancing the quality of financial risk datasets, we employ the widely used precision, Recall, F1, and the following side-evaluation metrics:

\textbf{Area Under the ROC Curve (AUC)}: The AUC metric is a key indicator of a classifier's overall performance. It represents the area under the ROC curve, mapping the true positive rate (TPR) against the false positive rate (FPR) at various thresholds. AUC ranges from 0 to 1, with higher values indicating better predictive performance. The AUC is formally expressed as:
\begin{equation}
    AUC = \int_{0}^{1} TPR(FPR^{-1}(x)) \, dx
\end{equation}

\textbf{Accuracy (Acc)}: This metric calculates the proportion of accurately predicted observations to the total observations. Notwithstanding, its applicability diminishes in imbalanced datasets, potentially inflating performance metrics due to the majority class's dominance:
\begin{equation}
    Accuracy = \frac{TP + TN}{TP + TN + FP + FN}
\end{equation}


\textbf{Kolmogorov-Smirnov (KS)}: The KS statistic measures the maximum difference between the cumulative distributions of true positive and false positive rates, indicating the model's ability to distinguish between positive and negative cases. Higher KS values imply better discrimination:
\begin{equation}
    KS = \max (TPR - FPR)
\end{equation}

\begin{figure*}[ht]
    \centering
    \includegraphics[width=0.9\textwidth]{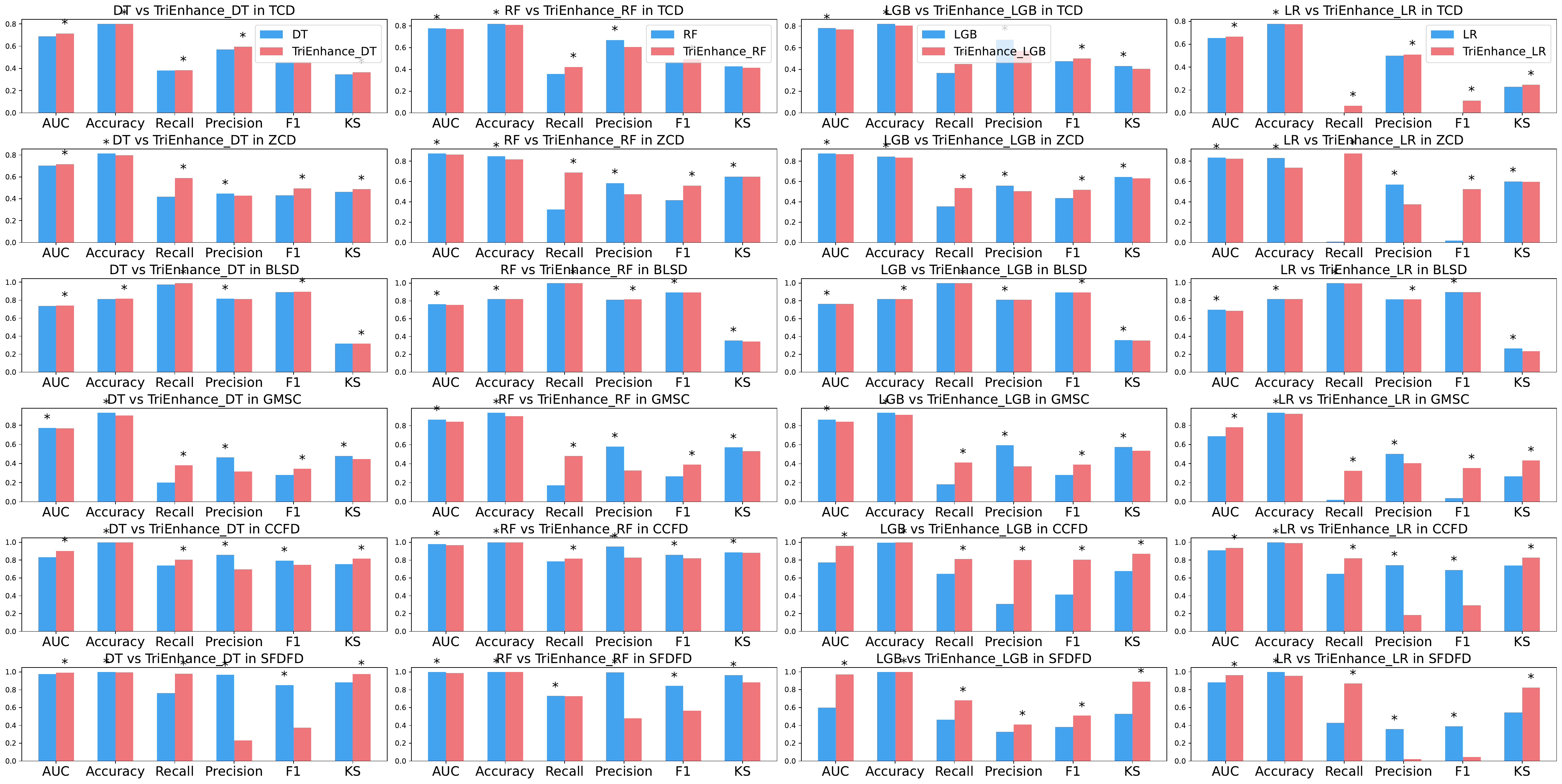}
    \caption{Comparison of main experimental results across datasets. * Indicates the higher value in each group.}
    \label{fig:Comparison of main experimental results across datasets}
    \vspace{-0.2cm}
\end{figure*}

\subsection{Implementation}
\textbf{Experimental Environment.} We conducted the experiments within a controlled hardware and software environment to ensure result accuracy and reproducibility. The hardware used consisted of an Intel(R) Xeon(R) Silver 4214R CPU @ 2.40GHz, equipped with 12 cores, accelerated by one piece NVIDIA GeForece-RTX 3080Ti. 

\textbf{Model Settings.} We utilized a stratified three-fold cross-validation approach, ensuring a consistent class ratio across training and validation sets. Baseline models, including Decision Tree(DT), Random Forest(RF), LightGBM(LGB), and Logistic Regression(LR). DT and RF were set with ${\text{max\_depth}} = 12$, and RF had ${\text{n\_estimators}} = 50$. LGB parameters were optimized to ${\text{n\_estimators}} = 50$ and ${\text{max\_depth}} = 12$. LR was implemented with its default settings. A consistent random seed of $42$ was utilized throughout all model training to ensure the reproducibility of the experiments. For reproducibility, source codes are uploaded to https://anonymous.4open.science/r/TriEnhance.

\subsection{Main Results}
We exploit four classic machine learning methods to evaluate TriEnhance, including Decision Tree(DT), Random Forest(RF), LightGBM(LGB), and Logistic Regression(LR). Model performance can be found in Figure~\ref{fig:Comparison of main experimental results across datasets}, along with the table version in the Appendix.

In general, models trained on datasets enhanced by TriEnhance exhibit significant improvements in AUC, recall, and F1 scores compared to datasets before enhancement with TriEnhance, demonstrating the effective enhancement of the original data environment. However, in highly imbalanced datasets such as CCFD and SFDFD, the utilization of TriEnhance for dataset enhancement can cause models to become overly sensitive to minority class samples, particularly in logistic regression models. This heightened sensitivity is believed to result from the models becoming excessively responsive to outlier samples during training on the enhanced data. While this increased sensitivity enhances the recall rate, it may also lead to a rise in false positives (FP). For example, in CCFD, where the imbalance ratio is as high as $1:587.65$, it indicates that for every $587$ legitimate transaction, there is one fraudulent transaction. In such scenarios, even a small number of misclassified legitimate transactions (FP) could surpass the actual fraud cases (positives). Since precision is calculated as the ratio of true positives (TP) to the sum of true positives and false positives (FP), even a slight increase in misclassified legitimate transactions can significantly and adversely affect precision, consequently resulting in a notable decline in the F1 score.

\begin{table}[h]
    \vspace{-0.2cm}
    \caption{Ablation study of TriEnhance in ZCD}
    \label{table: ablation_study_combined_ZCD}
    \centering
    \footnotesize
    \resizebox{\columnwidth}{!}{
    \begin{tabular}{llccc} 
    \toprule
    Model & Configuration & AUC & Recall & F1 \\
    \midrule
    \multirow{4}{*}{DT} 
    & {\bf TriEnhance} & {\bf 0.7177$\pm$0.0312} & {\bf 0.5882$\pm$0.0158} & {\bf 0.4965$\pm$0.0137} \\
    & \quad w/o sl & 0.6906$\pm$0.0328 & 0.5716$\pm$0.0189 & 0.4767$\pm$0.0102 \\
    & \quad w/o fil & 0.7145$\pm$0.0278 & 0.5847$\pm$0.0356 & 0.4953$\pm$0.0192 \\
    & \qquad w. KFULF & 0.7071$\pm$0.0121 & 0.5924$\pm$0.0345 & 0.4992$\pm$0.0171 \\
    & \qquad w. DDS & 0.6948$\pm$0.0251 & 0.5728$\pm$0.0242 & 0.4795$\pm$0.0167 \\
    & \quad w/o sl+fil & 0.6993$\pm$0.0320 & 0.5716$\pm$0.0054 & 0.4785$\pm$0.0040 \\
    \midrule
    \multirow{4}{*}{RF} 
    & {\bf TriEnhance} & {\bf 0.8667$\pm$0.0082} & {\bf 0.6880$\pm$0.0380} & {\bf 0.5606$\pm$0.0157} \\
    & \quad w/o sl & 0.8650$\pm$0.0093 & 0.6851$\pm$0.0240 & 0.5587$\pm$0.0138 \\
    & \quad w/o fil & 0.8640$\pm$0.0091 & 0.6673$\pm$0.0302 & 0.5518$\pm$0.0118 \\
    & \qquad w. KFULF & 0.8655$\pm$0.0096 & 0.6899$\pm$0.0307 & 0.5577$\pm$0.0143 \\
    & \qquad w. DDS & 0.8638$\pm$0.0088 & 0.6833$\pm$0.0375 & 0.5559$\pm$0.0197 \\
    & \quad w/o sl+fil & 0.8640$\pm$0.0091 & 0.6673$\pm$0.0302 & 0.5518$\pm$0.0118 \\
    \midrule
    \multirow{4}{*}{LGB} 
    & {\bf TriEnhance} & {\bf 0.8674$\pm$0.0059} & {\bf 0.5348$\pm$0.0264} & {\bf 0.5188$\pm$0.0134} \\
    & \quad w/o sl & 0.8647$\pm$0.0043 & 0.5336$\pm$0.0082 & 0.5163$\pm$0.0089 \\
    & \quad w/o fil & 0.8662$\pm$0.0048 & 0.5365$\pm$0.0170 & 0.5171$\pm$0.0078 \\
    & \qquad w. KFULF & 0.8674$\pm$0.0039 & 0.5294$\pm$0.0251 & 0.5167$\pm$0.0065 \\
    & \qquad w. DDS & 0.8654$\pm$0.0048 & 0.5175$\pm$0.0247 & 0.5066$\pm$0.0105 \\
    & \quad w/o sl+fil & 0.8652$\pm$0.0032 & 0.5318$\pm$0.0113 & 0.5145$\pm$0.0051 \\
    \midrule
    \multirow{4}{*}{LR} 
    & {\bf TriEnhance} & {\bf 0.8237$\pm$0.0087} & {\bf 0.8746$\pm$0.0321} & {\bf 0.5256$\pm$0.0130} \\
    & \quad w/o sl & 0.8296$\pm$0.0008 & 0.8283$\pm$0.0692 & 0.5214$\pm$0.0156 \\
    & \quad w/o fil & 0.8235$\pm$0.0083 & 0.8705$\pm$0.0284 & 0.5245$\pm$0.0123 \\
    & \qquad w. KFULF & 0.8244$\pm$0.0071 & 0.8378$\pm$0.0849 & 0.5192$\pm$0.0214 \\
    & \qquad w. DDS & 0.8200$\pm$0.0119 & 0.7350$\pm$0.2642 & 0.4830$\pm$0.0635 \\
    & \quad w/o sl+fil & 0.8272$\pm$0.0109 & 0.7296$\pm$0.1870 & 0.5010$\pm$0.0392 \\
    \bottomrule
    \end{tabular}}
    \vspace{-0.2cm}
\end{table}

\subsection{Ablation study}
We now investigate the impact of each proposed block in TriEnhance. The experiment results presented in Table~\ref{table: ablation_study_combined_ZCD}, are based on the Zhongyuan Credit Dataset (ZCD). We correspondingly examine three variants of Trienhance, including: ``w/o sl+fil" removes the data filtering block and self-learning block; ``w/o sl" removes the self-learning block; and ``w/o fil" removes the data filtering block. Specifically, we present two extra variants on model ``w/o fil" as: ``w/o fil w. KFULF" only reserves the KFULF strategy in the self-learning block, and ``w/o fil w. DDS" only reserves the DDS strategy in the self-learning block. Our findings reveal that (1) removing any of these components leads to a degradation in model performance, underscoring the positive contribution of all blocks; (2) eliminating the self-learning block and data filtering block results in the most significant performance decline, highlighting their complementary nature; (3) the proposed KFULF strategy performs well than DDS on four classic models, demonstrating its superiority.

\section{Conclusion}
In this study, we introduced TriEnhance, which offers an effective approach to enhancing the quality of imbalanced financial risk datasets through data synthesis, dynamic data filtering, and self-learning strategies. Within the self-learning block of TriEnhance, we introduced the innovative KFULF algorithm, an incremental enhancement method capable of continuously extracting valuable information from unlabeled data. Extensive experiments on various financial risk datasets illustrate that TriEnhance notably enhances the data environment, improving the ability of existing models to detect minority events. However, we observed that in highly imbalanced datasets, the model may demonstrate over-sensitivity to minority classes. Moving forward, we aim to further refine TriEnhance to enhance its capability in handling extremely imbalanced datasets, while also evaluating its compatibility with other deep learning models. 









\end{document}